\documentclass[runningheads]{llncs}
\usepackage{graphicx}
\usepackage{booktabs}
\usepackage[colorinlistoftodos]{todonotes}
\usepackage[inline]{enumitem}

\usepackage{chngcntr}
\counterwithout{figure}{chapter}
\counterwithout{table}{chapter}

\begin{document}
\title{User-friendly Comparison of Similarity Algorithms on Wikidata}

\author{Filip Ilievski \and Pedro Szekely \and Gleb Satyukov \and Amandeep Singh}
\authorrunning{Ilievski et al.}
%
\institute{Information Sciences Institute, University of Southern California \\ \email{\{ilievski,pszekely,gleb,amandeep\}@isi.edu} }
\maketitle              
\setcounter{footnote}{0}
\setcounter{table}{0}

\begin{abstract}\let\thefootnote\relax\footnotetext{Copyright © 2021 for this paper by its authors. Use permitted under Creative Commons License Attribution 4.0 International (CC BY 4.0).}
While the similarity between two concept words has been evaluated and studied for decades, much less attention has been devoted to algorithms that can compute the similarity of nodes in very large knowledge graphs, like Wikidata. To facilitate investigations and head-to-head comparisons of similarity algorithms on Wikidata, we present a user-friendly interface that allows flexible computation of similarity between Qnodes in Wikidata. At present, the similarity interface supports four algorithms, based on: graph embeddings (TransE, ComplEx), text embeddings (BERT), and class-based similarity. We demonstrate the behavior of the algorithms on representative examples about semantically similar, related, and entirely unrelated entity pairs. To support anticipated applications that require efficient similarity computations, like entity linking and recommendation, we also provide a REST API that can compute most similar neighbors for any Qnode in Wikidata.

\end{abstract}

\section{Introduction}

While the similarity between two concept words have been evaluated and studied for decades, much less attention has been devoted to algorithms that can compute similarity of nodes in very large knowledge graphs, like Wikidata. Effective and efficient metrics of Wikidata similarity are essential for a range of downstream applications, such as entity linking~\cite{cetoli2019neural,delpeuch2019opentapioca} and recommendation~\cite{gleim2020schematree,alghamdi2021learning}.

To facilitate investigations and head-to-head comparisons of similarity algorithms on Wikidata, we present a user-friendly graphical user interface (GUI) that allows flexible computation of similarity between Qnodes in Wikidata. The similarity interface is publicly available at \url{https://kgtk.isi.edu/similarity}. At present, the similarity GUI supports four algorithms, based on: graph embeddings(TransE~\cite{bordes2013translating}, ComplEx~\cite{trouillon2016complex}), text embeddings (BERT~\cite{devlin2018bert}), and class-based similarity.
Through the similarity interface, users can investigate the ability of different (families of) algorithms to capture similarity of concepts and entities in Wikidata. To support applications that require efficient similarity computations, like entity linking and recommendation, we also provide a REST API that can compute the most similar neighbors for any Qnode in Wikidata. The endpoint of this API is \url{https://dsbox02.isi.edu:8888/nearest-neighbors}.

We demonstrate the behavior of the algorithms on representative examples about semantically similar, related, or entirely unrelated entity pairs. We show that the class-based metric consistently captures semantic similarity, and assigns lower scores to terms that are merely related or unrelated. BERT-based similarity behaves differently, providing high scores to both semantically similar and related pairs. The graph embedding-based metrics are somewhere in between class-based similarity and BERT.


The code for the similarity GUI and our similarity API is freely available on GitHub: \url{https://github.com/usc-isi-i2/kgtk-similarity}.

\section{Similarity interfaces}

In this section, we describe the similarity interfaces that we have developed, together with their currently supported algorithms.

\noindent \textbf{GUI} Our GUI allows users to search for a primary Qnode based on its labels or aliases. The user could then add any number of secondary Qnodes in the same way, based on free text search against the node labels and aliases. We use ElasticSearch to build a text index and enable this search. The interface then displays the similarity between the primary node and each secondary Qnode, according to each of the supported algorithms. 

Currently, we support four algorithms:

\begin{enumerate}
    \item \textbf{Class similarity} computes the set of common \textit{is-a} parents for two nodes. Here, the is-a relations are computed as a transitive closure over both the subclass-of (P279) and the instance-of (P31) relations. Each shared parents is weighted by its inverse document frequency (IDF), computed based on the number of instances that transitively belong to that parent class.
    \item \textbf{TransE similarity} computes the cosine similarity between the TransE embeddings of two Wikidata nodes. 
    \item \textbf{ComplEx similarity} computes the cosine similarity between the TransE embeddings of two Wikidata nodes. 
    \item \textbf{Text similarity} computes the cosine similarity between the BERT embeddings of two Wikidata nodes. We pre-compute these BERT embeddings over a lexicalized version of each Wikidata Qnode, based on its outgoing edges in the graph.  
\end{enumerate}

In practice, we use the operation \texttt{graph-embeddings} of the Knowledge Graph ToolKit (KGTK)~\cite{ilievski2020kgtk} to compute TransE and ComplEx embeddings. We use the KGTK \texttt{text-embeddings} command to compute the text (BERT-based) embeddings.
A snapshot of the similarity interface is shown in Figure~\ref{fig:motors}.

\noindent \textbf{Nearest Neighbors API} Our REST API returns $K$ nearest neighbors for a Qnode based on the ComplEx algorithm. We index the ComplEx embeddings in a FAISS~\cite{JDH17} index, which facilitates efficient retrieval.

\section{Analysis}

\textbf{GUI examples}
In this section, we show the similarity scores provided by the supported algorithms between the Wikidata Qnode for motorcycle (Q34493) and ten other Qnodes. Specifically, we include three semantically similar nodes: bus (Q5638), Dirt Bike (Q3050907), and yacht (Q170173); four related, but dissimilar nodes: engine (Q44167), helmet (Q173603), road (Q34442), and cyclist (Q2125610); and three unrelated nodes: cheese (Q10943), Norway (Q20), and shelf (Q2637814). Following our terminology introduced in the previous section, motorcycle is the primary Qnode, and the ten additional Qnodes are secondary.

\begin{figure}
    \centering
    \includegraphics[width=\textwidth]{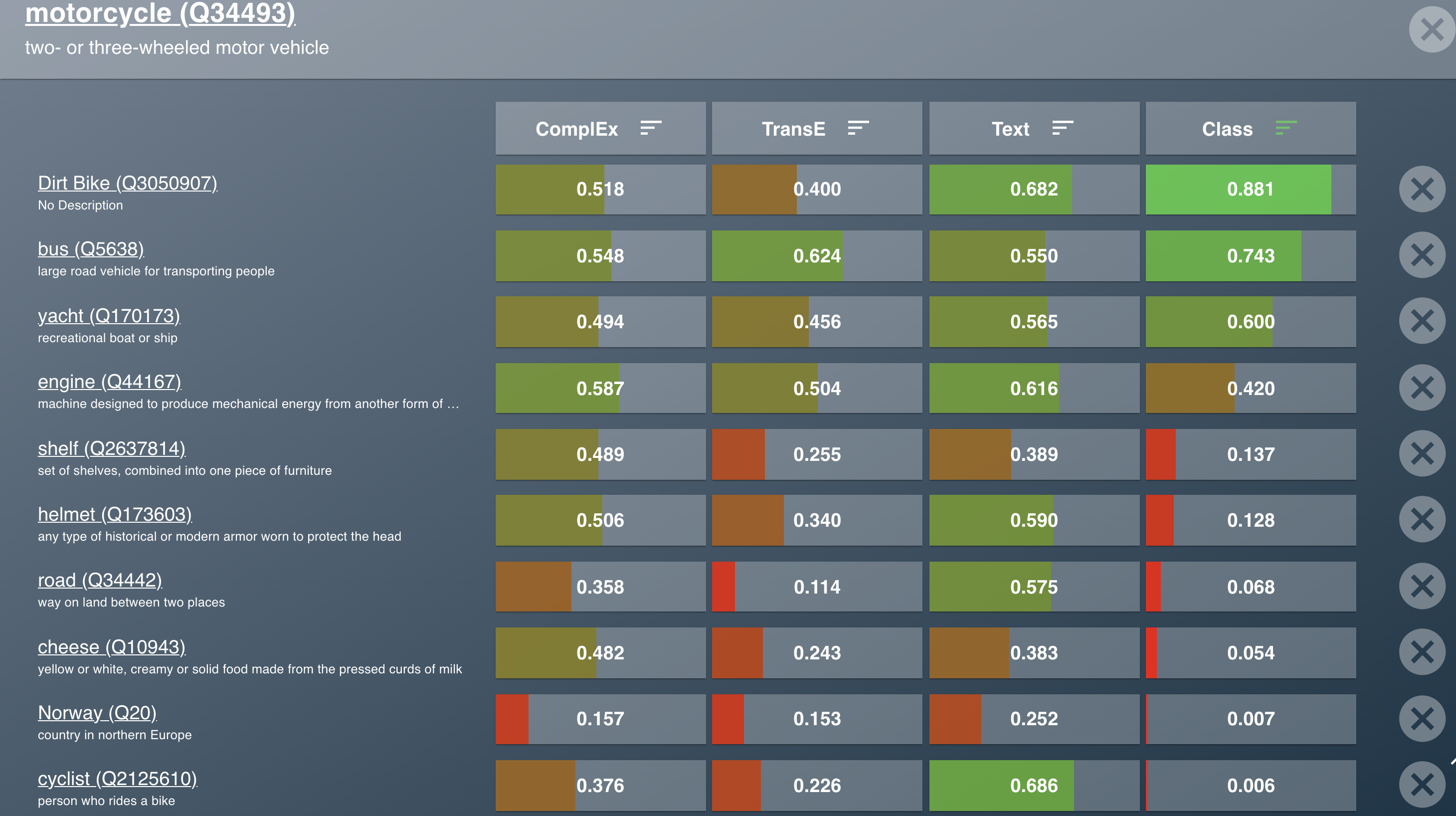}
    \caption{Similarity between motorcycle (Q34493) and ten other terms, i.e., three semantically similar nodes: bus (Q5638), Dirt Bike (Q3050907), and yacht (Q170173); four related, but dissimilar nodes: engine (Q44167), helmet (Q173603), road (Q34442), and cyclist (Q2125610); and three unrelated nodes: cheese (Q10943), Norway (Q20), and shelf (Q2637814). The results are ordered based on their \textit{class similarity score}.}
    \label{fig:motors}
\end{figure}

Table~\ref{fig:motors} shows the obtained similarity scores, in a descending order according to their class-based scores. We observe that the class-based metric consistently prioritizes semantically similar nodes over the others, as its three top-scored nodes are semantically similar to motorcycle. The remaining nodes receive notably lower scores, with the exception of the motorcycle-engine pair, whose similarity is fairly high (0.42). We observe that the class metric makes little distinction between nodes that are related and nodes that are unrelated to motorcycle. These findings show that the class metric mostly captures semantic similarity, and it does not capture semantic relatedness. This is intuitive, given that it is purely based on the Wikidata \textit{taxonomy}, and naturally favors semantically similar terms.

\begin{figure}
    \centering
    \includegraphics[width=\textwidth]{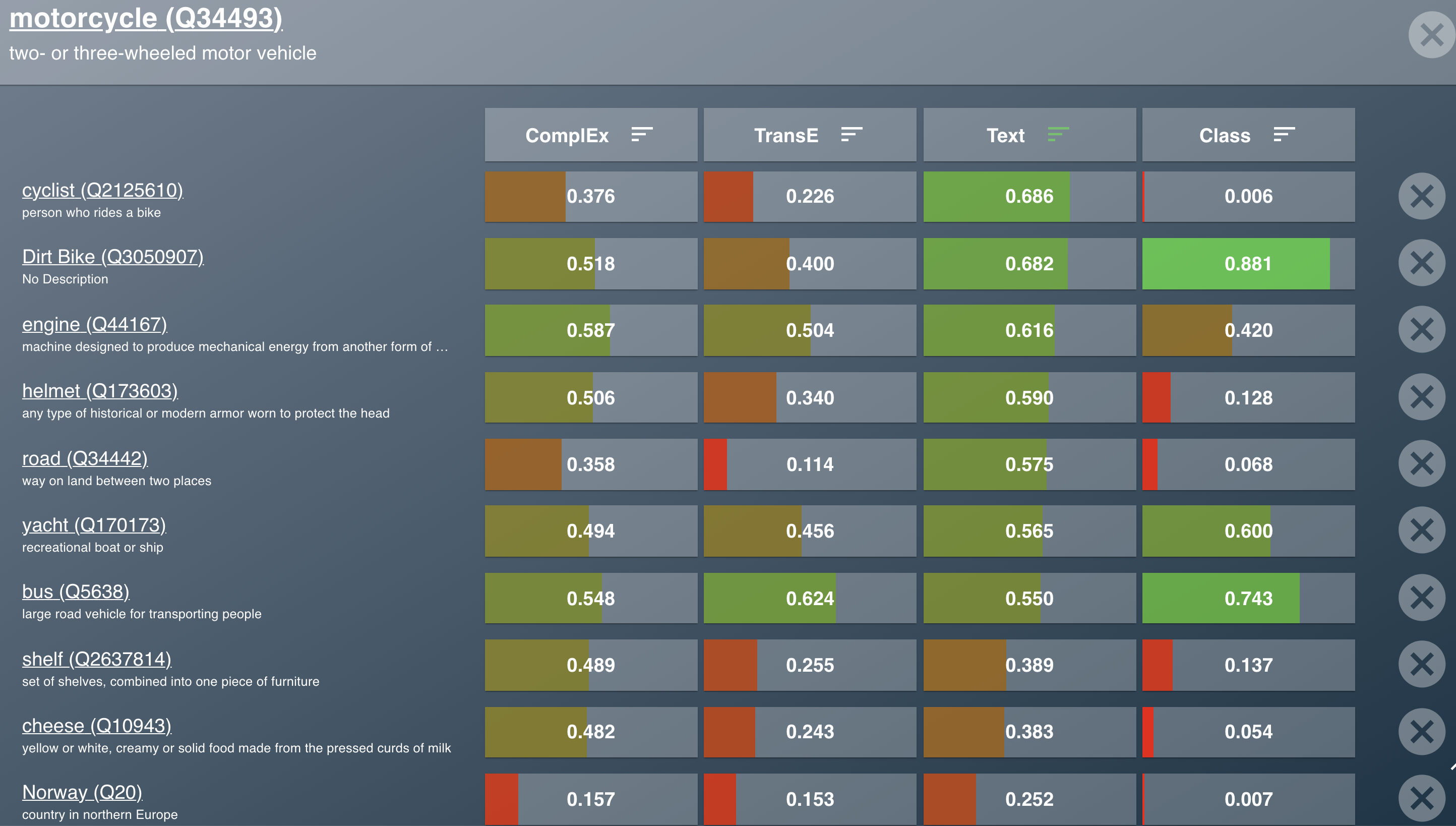}
    \caption{Similarity between motorcycle (Q34493) and ten other terms, i.e., three semantically similar nodes: bus (Q5638), Dirt Bike (Q3050907), and yacht (Q170173); four related, but dissimilar nodes: engine (Q44167), helmet (Q173603), road (Q34442), and cyclist (Q2125610); and three unrelated nodes: cheese (Q10943), Norway (Q20), and shelf (Q2637814). The results are ordered based on their \textit{text similarity score}.}
    \label{fig:motors_text}
\end{figure}

Next, we order the same set of results based on their text-based score. The result is shown in Figure~\ref{fig:motors_text}. Here, we observe that the terms that are unrelated to motorcycle (shelf, cheese, and Norway) are consistently assigned low scores. At the same time, we observe that the terms that are semantically similar (e.g., dirt bike) and merely related (e.g., cyclist) receive comparable scores. We conclude that the BERT-based text similarity metric is able to discern related from unrelated nodes, but it is unable to distinguish between similar and related terms. This can be expected, considering that the BERT model is trained to capture \textit{natural language co-occurrence}, thus favoring both semantic and related terms over unrelated ones.



\begin{figure}
    \centering
    \includegraphics[width=\textwidth]{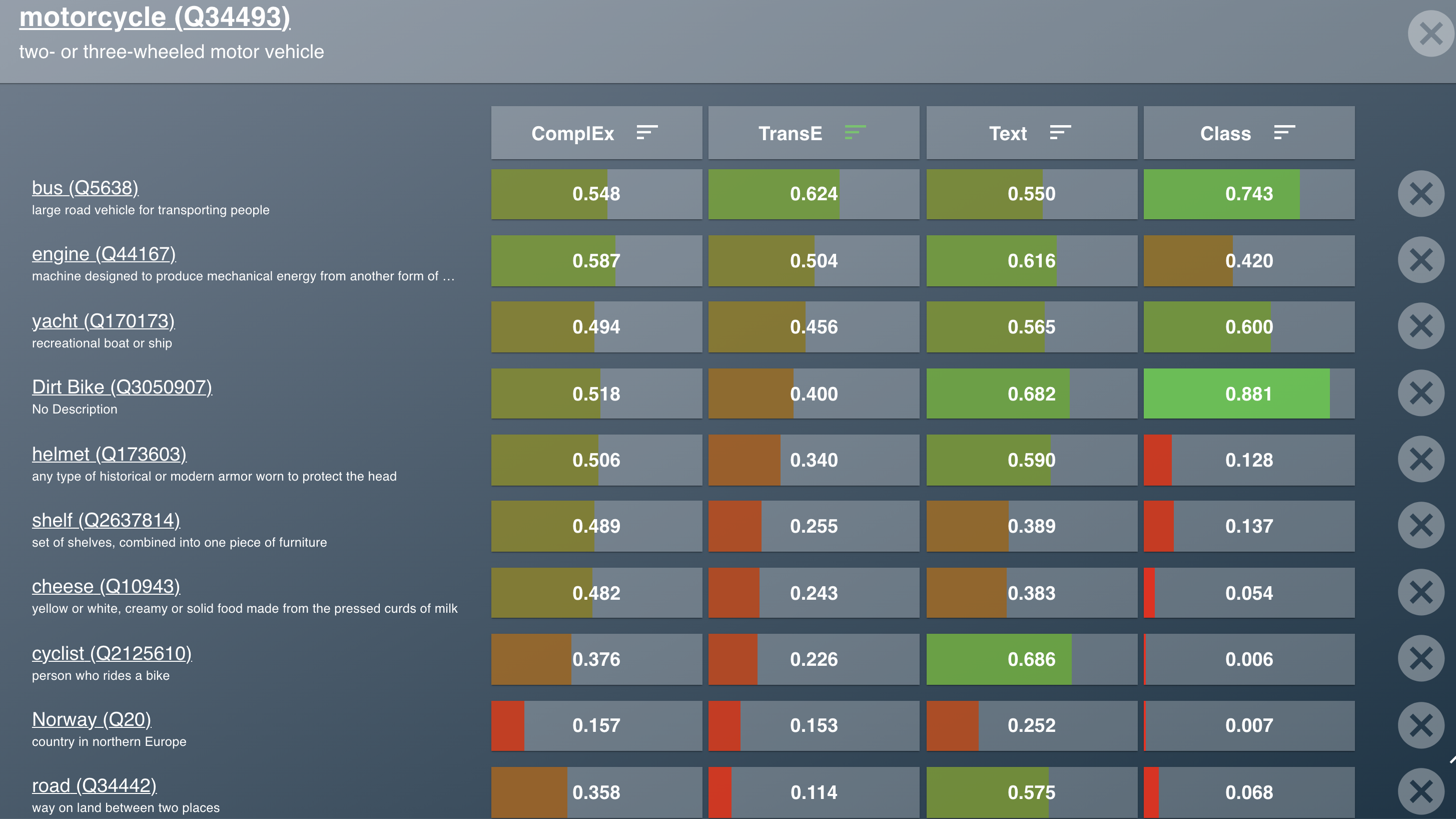}
    \caption{Similarity between motorcycle (Q34493) and ten other terms, i.e., three semantically similar nodes: bus (Q5638), Dirt Bike (Q3050907), and yacht (Q170173); four related, but dissimilar nodes: engine (Q44167), helmet (Q173603), road (Q34442), and cyclist (Q2125610); and three unrelated nodes: cheese (Q10943), Norway (Q20), and shelf (Q2637814). The results are ordered based on their \textit{TransE score}.}
    \label{fig:motors_trans}
\end{figure}

Figure~\ref{fig:motors_trans} provides a third ordering of the results, based on their TransE score. The scoring in this case correlates to a lesser extent with our a priori three-way categorization of the Qnodes, though on average semantic similarity is favored over relatedness, which is on average favored over unrelatedness. This could be explained with the property of the graph embeddings to capture \textit{structural} similarity of nodes, i.e., to assign higher similarity between nodes that connect to similar other nodes (e.g., both engine and bus relate to car). For this reason, engine, bus, and helmet are assigned higher similarity than terms such as Norway and road.

\noindent \textbf{Nearest neighbors API examples} The nearest neighbors API can be leveraged to obtain the top-K most similar Wikidata Qnodes for a given Qnode. For instance, in order to obtain the top-5 most similar nodes to motorcycle (Q34493), we query: \url{https://dsbox02.isi.edu:8888/nearest-neighbors?qnode=Q34493\&k=5}.
The result is a list of the 5 most similar nodes, with their corresponding distance from the motorcycle Qnode and their human-readable label in English:

\begin{small}
\begin{verbatim}
[
    {
        qnode: "Q13586807",
        score: 13.393990516662598,
        label: "Manet Korado"
    },
    {
        qnode: "Q376498",
        score: 13.482695579528809,
        label: "diesel motorcycle"
    },
    {
        qnode: "Q28126796",
        score: 15.886520385742188,
        label: "Harley-Davidson FLSTFB Fat Boy"
    },
    {
        qnode: "Q20076361",
        score: 16.452970504760742,
        label: "Honda SH50"
    },
    {
        qnode: "Q18695780",
        score: 16.553009033203125,
        label: "Bultaco TSS Mk2"
    }
]
\end{verbatim}
\end{small}

Curiously, the list of the most similar nodes is dominated by specific motorcycle models and categories. The most similar three nodes are direct subclasses of the motorcycle class (connected by using the P279 relation). The remaining two Qnodes are specific models of motorcycles, represented as instances (P31) of the motorcycle class in Wikidata. This confirms our earlier observation: the graph embeddings like ComplEx assign a higher similarity to node pairs that connect to similar structures in the Wikidata graph.

\section{Similarity in Downstream Tasks}

Meaningful estimation of similarity is at the core of a long list of applications in natural language processing, information retrieval, and network analysis. Here, we discuss the role of estimating similarity for two prominent applications: 1) entity linking in tables, and 2) recommendation and deduplication. We also discuss how our interfaces could support these applications.

\textbf{Entity linking in tables} Understanding the reference of entities in tables relies on two different notions of similarity. On the one hand, entities in the same column typically are of the same type, or play the same role in a given context. For example, a table with Russian politicians will include a column with politicians, and a column with their positions. Thus, understanding entities within a column relies on similarity indicators that can capture semantic similarity, such as our class-based metric. On the other hand, entities mentioned in the same row rely on metrics that capture aspects of relatedness, such as our text-based metric, which relies on linguistic similarity, or our graph embedding metrics, which capture structural similarity. Following our previous example, this would require a metric that can assign a high score to the pairs: Vladimir Putin - Russia, and Vladimir Putin - president.

\textbf{Recommendation and deduplication} 
A special use case of Qnode recommendation is assistance of Wikidata editors. Namely, when an editor introduces a new Qnode, it is useful to have metrics which can detect very similar existing entities and ask the editor to confirm that the new entity is different from the most similar existing ones~\cite{alghamdi2021learning}. This procedure would help to avoid introducing duplicates in Wikidata, which is a key challenge today, considering that millions of redirects have been introduced in Wikidata since its inception~\cite{shenoy2021study}. At the same time, similarity methods could be run over the current set of entities in Wikidata to detect potentially existing duplicates, which can be validated by an editor before their merging. The class-based metric could be used to detect potential duplicates, and it could be complemented with additional metrics (e.g., text-based similarity) when the taxonomic information is not present for a node.



\section{Conclusions}

This demo paper presented a user-friendly interface for computation of pairwise similarity between Qnodes in Wikidata. To facilitate head-to-head comparisons of similarity, the interface rendered the scores for multiple node pairs by four different algorithms: a class-based metric, two graph embedding metrics, and a language model based (text) metric. We experimented with their scores on semantically similar, related, or entirely unrelated entity pairs, observing that the class-based metric favored semantically similar pairs, while the text-based metric favored both semantically similar and related pairs, at the expense of the unrelated ones. Graph embeddings scored pairs orthogonally to our similarity categorization, by assigning higher scores to pairs that are structurally similar in Wikidata. To support  applications where similarity plays a key role, such as entity linking, recommendation, and deduplication, we also provided a public API that returns the top-K neighbors for a given Qnode.


\bibliographystyle{splncs04}
\bibliography{refs}

\end{document}